\documentclass[11pt,a4paper]{article}
\usepackage[hyperref]{emnlp-ijcnlp-2019}
\usepackage{times}
\usepackage{latexsym}
\usepackage{graphicx}

\usepackage{hyperref}
\usepackage{url}

\usepackage{amsmath}
\usepackage{algorithm}
\usepackage{verbatim}
\usepackage[noend]{algpseudocode}

\aclfinalcopy 

\title{Fine-Tuned Neural Models for Propaganda Detection at the Sentence and Fragment levels}

\author{Tariq Alhindi$^\dag$ \hspace{2cm}
        Jonas Pfeiffer$^*$ \hspace{2cm}
        Smaranda Muresan$^{\dag \ddag}$\\
  $^\dag$Department of Computer Science, Columbia University\\
  $^\ddag$Data Science Institute, Columbia University\\
  $^*$Ubiquitous Knowledge Processing Lab, Technische Universitat Darmstadt \\
  {\tt \{tariq.a, smara\}@columbia.edu} \\
  {\tt pfeiffer@ukp.informatik.tu-darmstadt.de}}

\date{}

\begin{document}

\maketitle
\begin{abstract}
 This paper presents the CUNLP submission for the NLP4IF 2019 shared-task on Fine-Grained Propaganda Detection. Our system finished 5th out of 26 teams on the sentence-level classification task and 5th out of 11 teams on the fragment-level classification task based on our scores on the blind test set. We present our models, a discussion of our ablation studies and experiments, 
 and an analysis of our performance on all eighteen propaganda techniques present in the corpus of the shared task.
\end{abstract}



\section{Introduction}
Propaganda aims at influencing a target audience with a specific group agenda using faulty reasoning and/or emotional appeals \cite{miller1939}. Automatic detection of propaganda has been studied mainly at the article level \cite{rashkin2017truth, barron2019proppy}. However, in order to build computational models that can explain why an article is propagandistic, the model would need to detect specific techniques present at sentence or even token level. 

The NLP4IF shared task on fine-grained propaganda detection aims to produce models capable of spotting propaganda techniques in sentences and text fragments in news articles \cite{NLP4IF2019:propaganda:task}. The data for this task consist of news articles that were labeled at the fragment level with one of eighteen propaganda techniques. 

There are two sub-tasks in this shared task. The first one is a sentence classification task (SLC) to detect whether a sentence has a propaganda fragment or not. This binary classification task is evaluated based on the F1 score of the propaganda class which approximately represents one-third of the data. The second sub-task is a fragment level classification  (FLC) task, in which a system needs to detect the type of propaganda technique expressed in a text fragment together with the beginning and the end of that text fragment. 
This task is evaluated based on the prediction of the type of propaganda technique and the intersection between the gold and the predicted spans. 
The details to the evaluation measure used for the FLC task are explained in \citet{NLP4IF2019:propaganda:task}.
Both sub-tasks were automatically evaluated on a unified development set. 
The system performance was centrally assessed without distributing the gold labels, however allowing for an unlimited number of submissions.
The final performance on the test set was similarly evaluated, with the difference that the feedback was given only after the submission was closed, simultaneously concluding the shared-task.


In this paper, we describe the data in Section \ref{sec:data}, our proposed methods for both sub-tasks in Section \ref{sec:methods}, and analyze the results and errors of our models in Section \ref{sec:results}.

\section{Data}
\label{sec:data}
The data for this shared task includes 350 articles in the training set, 61 articles in the development set, and 86 articles in the test set. The articles were taken from 48 news outlets; 13 propagandistic and 35 non-propagandistic as labeled by Media Bias/Fact Check\footnote{\href{https://mediabiasfactcheck.com/}{https://mediabiasfactcheck.com/}}. These articles were annotated at the fragment level where each annotator was asked to tag the start and end of the propaganda text span as well as the type of propaganda technique. Table \ref{table:propaganda_techniques} lists all eighteen propaganda techniques and their frequencies in the training data. Since submissions to the development set were closed after the release of the test set, we divided the training set (350 articles) into a training set of 280 articles and a local dev set of 70 articles to continue to be able to perform ablation studies. We also conduct our error analysis on the local dev set because we do not have access to the gold labels of the official dev and test sets of the shared task.

\begin{table}[]
\center
\scalebox{0.77}{
\begin{tabular}{l l}
\hline \hline
\textbf{Propaganda Technique} & \textbf{Frequency} \\ \hline
Loaded Language & 2,115 \\
Name Calling,Labeling & 1,085 \\
Repetition & 571 \\
Doubt & 490 \\
Exaggeration,Minimisation & 479 \\
Flag-Waving & 240 \\
Appeal to Fear/Prejudice & 239 \\
Causal Oversimplification & 201 \\
Slogans & 136 \\
Appeal to Authority & 116 \\
Black-and-White Fallacy & 109 \\
Thought-terminating Cliches & 79 \\
Whataboutism & 57 \\
Reductio ad hitlerum & 54 \\
Red Herring & 33 \\
Bandwagon & 13 \\
Straw Men & 13 \\
Obfuscation,Intentional Vagueness,Confusion & 11 \\ \hline
\textbf{Total} & \textbf{6,041} \\
\hline \hline

\end{tabular}}
\caption{Frequency of all eighteen propaganda techniques in the training data}
\label{table:propaganda_techniques}
\end{table}

More details about the dataset and the annotation scheme for the eighteen propaganda techniques can be found in \citet{EMNLP19DaSanMartino}. However, the results on the shared task data are not directly comparable as more articles were added to shared task's data. \citet{NLP4IF2019:propaganda:task} should be referred to for an accurate comparison between participants who all used the same development and test sets.

\section{Methods}
\label{sec:methods}
In the following we explain the details of our approach for the SLC and FLC tasks. 
\subsection{Sentence Level Classification (SLC)}
We fine-tuned BERT \cite{devlin-etal-2019-bert} for the binary sentence-level classification task of {\tt propaganda} vs. {\tt non-propaganda}. The training set has 16,298 sentences, out of which 4,720 are from the propaganda class. We used {\tt bert-base-uncased} in our experiments as in preliminary results the cased version did not provide any improvements. The model was trained for 3 epochs using a learning rate of 2e-5, a maximum sequence length of 128, and a batch size of 16. We also experiment with a Logistic Regression Classifiers, where we used Linguistic Inquiry and Word Count (LIWC) features \cite{pennebaker2001linguistic}, punctuation features such as the existence of quotes or question marks, as well as BERT's prediction probabilities for each class. This gave some minor improvement on the development set of the shared-task. However, since we did not have access to the development set submission after the test set was released, we chose the final model based on the performance on the local development set. The final model used the fine-tuned BERT model mentioned above with a condition to predict {\tt non-propaganda} only if the prediction probability is above 0.70 for the non-propaganda class. Otherwise the prediction of the sentence will be {\tt propaganda} even if the majority of the prediction probability mass was for the {\tt non-propaganda} class. This was a way to handle the unbalance in the training data without having to discard part of the data. The 0.70 threshold was chosen after elaborate experiments on both the local and the shared-task's development sets. This condition consistently provided an improvement of around 5 points in F1 score of the propaganda class on all experiments using different sets of features as shown in Table \ref{table:SLC_results}.

\subsection{Fragment Level Classification (FLC)}
\label{sec:FLC}

Our architecture for the sequence labeling task builds on the flair framework \cite{akbik2018coling,akbik2019flair} that combines character level embeddings with different kinds of word embeddings as input to a BiLSTM-CRF model \cite{ma2016end,lample2016neural}. \citet{akbik2018coling} have shown that stacking multiple pre-trained embeddings as input to the LSTM improves performance on the downstream sequence labeling task.  We combine Glove embeddings \cite{pennington2014glove} with Urban Dictionary\footnote{\href{https://www.urbandictionary.com/}{https://www.urbandictionary.com/}} embeddings\footnote{\href{https://data.world/jaredfern/urban-dictionary-embedding}{https://data.world/jaredfern/urban-dictionary-embedding}}. 

Due to the small-size of our data set we additionally include one-hot-encoded features based on dictionary look-ups from the UBY dictionary provided by \citet{gurevych2012uby}. These features are based on concepts associated with the specific word such as \textit{offensive, vulgar, coarse,} or \textit{ethnic slur}. In total, 30 concept features were added as additional dimensions to the embedding representations.

We also experimented with stacking BERT embeddings with all or some of the embeddings mentioned above. However, this resulted on lower scores on both the local and shared task development sets. The best model used urban-glove embeddings with concatenated one-hot encoded UBY features stacked with both forward and backward flair embeddings. The model was trained for a maximum of 150 epochs with early stopping using a learning rate of 0.1, a batch size of 32, and a BiLSTM with hidden size 256. The results of this model are shown in Table \ref{table:flc_results}.

\section{Results and Error Analysis}
\label{sec:results}
In this section we discuss the results of both sub-tasks on all three datasets: the local development set, the shared task development and test sets.
\subsection{SLC Results}
In SLC, we ran multiple experiments using BERT with and without additional features as shown in Table \ref{table:SLC_results}. The features include using the text passed as is to BERT without any preprocessing. Also, we experimented with adding the context which includes the two sentences that come before and after the target sentence. Context sentences were concatenated and passed as the second BERT input, while the target sentence was passed as the first BERT input. In addition, we experimented with using BERT logits (i.e., the probability predictions per class) as features in a Logistic Regression (LR) classifier concatenated with handcrafted features (e.g., LIWC, quotes, questions), and with predictions of our FLC classifier (tagged spans: whether the sentence has a propaganda fragment or not). However, none of these features added any statistically significant improvements. Therefore, we used BERT predictions for our final model with a condition to predict the majority class {\tt non-propaganda} only if its prediction probability is more than 0.70 as shown in Table \ref{table:SLC_results_test}. This is a modified threshold as opposed to 0.80 in the experiments shown in Table \ref{table:SLC_results} to avoid overfitting on a one dataset. The final threshold of 0.70 was chosen after experiments on both the local and shared task development sets, which also represents the ratio of the {\tt non-propaganda} class in the training set.

\begin{table}[]
    \centering
    \scalebox{0.75}{
    \begin{tabular}{l | l | c c c}
    \hline \hline
        & & \multicolumn{3}{c}{\textbf{Development}} \\
        \textbf{Features} & \textbf{Model} & P & R & F \\
        \hline
        text & BERT & 0.69 & 0.55 & 0.61\\
        
        \textbf{text} & \textbf{BERT*} & \textbf{0.57} & \textbf{0.79} & \textbf{0.66}\\
        
        context & BERT & 0.70 & 0.53 & 0.60\\
        context & BERT* & 0.63 & 0.67 & 0.65\\
        
        BERT logits + handcrafted** & LR & 0.70 & 0.56 & 0.61\\
        BERT logits + handcrafted** & LR* & 0.60 & 0.71 & 0.65\\
        BERT logits + tagged spans & LR & 0.70 & 0.53 & 0.60\\
        BERT logits + tagged spans & LR* & 0.61 & 0.71 & 0.66\\
        BERT logits + all & LR & 0.71 & 0.52 & 0.60\\
        BERT logits + all & LR* & 0.61 & 0.71 & 0.66\\
        \hline \hline
    \end{tabular}}
    {\small \raggedright *Non-propaganda class is predicted only if its prediction \\ \hspace{0.1cm} probability is $>0.80$ \\ **handcrafted features include LIWC and presence of \\ \hspace{0.2cm} questions or quotes \par}
    \caption{SLC experiments on different feature sets}
    \label{table:SLC_results}
\end{table}

        

\begin{table}[]
    \centering
    \begin{tabular}{l| c c c}
    \hline \hline
         Dataset & P & R & F \\
         \hline
         Local Dev & 0.60 & 0.75 & 0.67 \\
         Development &  0.62 & 0.68 & 0.65\\
         Test & 0.58 & 0.66 & 0.618\\
         \hline \hline
    \end{tabular}
    
    {\small *Non-propaganda class is predicted only if its prediction \\ \hspace{0.1cm} probability is $>0.70$ \par}
    \caption{SLC best model results on all three datasets}
    \label{table:SLC_results_test}
\end{table}

\paragraph{Discussion of Propaganda Types:}
To further understand our model's performance in the SLC task, we looked at the accuracy of each propaganda techniques that occur more than 20 times in the local dev set as shown in Table \ref{table:SLC_errors}. {\tt Repetition} and {\tt Doubt} are the two most challenging types for the classifier even though they are in the four most frequent techniques. It is expected for {\tt Repetition} to be challenging as the classifier only looks at one sentence while {\tt Repetition} occurs if a word (or more) is repeatedly mentioned in the article. Therefore, more information needs to be given to the classifier such as word counts across the document of all words in a given sentence. Due to time constrains, we did not test the effect of adding such features. {\tt Doubt} on the other hand could have been challenging due to its very wide lexical coverage and variant sentence structure as doubt is expressed in many different words and forms in this corpus (e.g. ``How is it possible the pope signed this decree?" and ``I've seen little that has changed"). It is also among the types with high variance in length where one span sometimes go across multiple sentences.

\begin{table}[]
    \small
    \centering
    \begin{tabular}{l | l | l}
\hline \hline
\textbf{Technique} & \textbf{Count} & \textbf{Accuracy} \\
\hline
Loaded Language & 299 & 71\% \\
Name Calling,Labeling & 163 & 69\% \\
Repetition & 124 & 44\% \\
Doubt & 71 & 40\% \\
Exaggeration,Minimisation & 63 & 67\% \\
Flag-Waving & 35 & 74\% \\
Appeal to Fear/Prejudice & 42 & 52\% \\
Causal Oversimplification & 24 & 58\% \\
Slogans & 24 & 54\% \\
\hline \hline
    \end{tabular}
    \caption{SLC accuracy on frequent propaganda techniques in the local development set}
    \label{table:SLC_errors}
\end{table}

\subsection{FLC Results}
In FLC, we only show the results of our best model in Table \ref{table:flc_results} to focus more on the differences between propaganda techniques. A more elaborate study of performance of different models should follow in future work. The best model is a BiLSTM-CRF with flair and urban glove embeddings with one hot encoded features as mentioned in Section \ref{sec:FLC}. 
\paragraph{Discussion of Propaganda Types:}
As we can see in Table \ref{table:flc_results}, we can divide the propaganda techniques into three groups according to the model's performance on the development and test sets. The first group includes techniques with non-zero F1 scores on both datasets: {\tt Flag-Waving}, {\tt Loaded Language}, {\tt Name Calling,Labeling} and {\tt Slogans}. This group has techniques that appear frequently in the data and/or techniques with strong lexical signals (e.g. "American People" in {\tt Flag-Waving}) or punctuation signals (e.g. quotes in {\tt Slogans}). The second group has the techniques with a non-zero F1 score on only one of the datasets but not the other, such as: {\tt Appeal to Authority}, {\tt Appeal to Fear}, {\tt Doubt}, {\tt Reduction}, and {\tt Exaggeration}{\tt,Minimisation}. Two out of these five techniques ({\tt Appeal to Fear} and {\tt Doubt}) have very small non-zero F1 on the development set which indicates that they are generally challenging on our model and were only tagged due to minor differences between the two datasets. However, the remaining three types show significant drops from development to test sets or vice-versa. This requires further analysis to understand why the model was able to do well on one dataset but get zero on the other dataset, which we leave for future work. The third group has the remaining nine techniques were our sequence tagger fails to correctly tag any text span on either dataset. This group has the most infrequent types as well as types beyond the ability for our tagger to spot by looking at the sentence only such as {\tt Repetition}.
\paragraph{Precision and Recall:}
Overall, our model has the highest precision among all teams on both datasets, which could be due to adding the UBY one-hot encoded features that highlighted some strong signals for some propaganda types. This also could be the reason for our model to have the lowest recall among the top 7 teams on both datasets as having explicit handcrafted signals suffers from the usual sparseness that accompanies these kinds of representations which could have made the model more conservative in tagging text spans.

\begin{table}[]
\center
\scalebox{0.75}{
\begin{tabular}{l | c c c | c}
\hline \hline
\textbf{Propaganda} & \multicolumn{3}{c|}{\textbf{Development}} & \textbf{Test} \\
\textbf{Technique} & \textbf{P} & \textbf{R} & \textbf{F} & \textbf{F} \\
\hline
Appeal to Authority & 0 & 0 & 0 & 0.212\\ 
Appeal to Fear/Prejudice & 0.285 & 0.006 & 0.011 & 0\\
Bandwagon & 0 & 0 & 0 & 0\\
Black-and-White Fallacy & 0 & 0 & 0 & 0\\
Causal Oversimplification & 0 & 0 & 0 & 0\\
Doubt & 0.007 & 0.001 & 0.002 & 0\\
Exaggeration,Minimisation & 0.833 & 0.085 & 0.154 & 0\\
Flag-Waving & 0.534 & 0.102 & 0.171 & 0.195\\
Loaded Language & 0.471 & 0.160 & 0.237 & 0.130\\
Name Calling,Labeling & 0.270 & 0.112 & 0.158 & 0.150\\
O,IV,C & 0 & 0 & 0 & 0\\
Red Herring & 0 & 0 & 0 & 0\\
Reductio ad hitlerum & 0.318 & 0.069 & 0.113 & 0\\
Repetition & 0 & 0 & 0 & 0\\
Slogans & 0.221 & 0.034 & 0.059 & 0.003\\
Straw Men & 0 & 0 & 0 & 0\\
Thought-terminating Cliches & 0 & 0 & 0 & 0\\
Whataboutism & 0 & 0 & 0 & 0\\
\hline
\textbf{Overall} & 0.365 & 0.073 & 0.122 & 0.131$^{*}$\\
\hline \hline
\end{tabular}}
{ \small \raggedright 
      *Test set overall precision is 0.323 and recall is 0.082. \\ \hspace{0.1cm} Precision and recall per technique were not provided for \\ \hspace{0.1cm} the test set by the task organizers. \par}
\caption{Precision, recall and F1 scores of the FLC task on the development and test sets of the shared task.}
\label{table:flc_results}
\end{table}

\subsection{Remarks from Both Tasks}
In light of our results on both sub-tasks, we notice that the BERT-based sentence classification model is performing well on some propaganda types such as {\tt Loaded Language} and {\tt Flag-Waving}. It would be interesting to test in future work if using BERT as a sequence tagger (and not BERT embeddings in a BiLSTM-CRF tagger like we tested) would help in improving the sequence tagging results on those particular types. Finally, we noticed two types of noise in the data; there were some duplicate articles, and in some articles the ads were crawled as part of the article and tagged as non-propaganda. These could have caused some errors in predictions and therefore investigating ways to further clean the data might be helpful. 

\section{Conclusion}
Propaganda still remains challenging to detect with high precision at a fine-grained level. This task provided an opportunity to develop computational models that can detect propaganda techniques at sentence and fragment level. We presented our models for each sub-task and discussed challenges and limitations. For some propaganda techniques, it is not enough to only look at one sentence to make an accurate prediction (e.g. {\tt Repetition}) and therefore including the whole article as context is needed. For future work, we want to experiment with using a BERT-based sequence tagger for the FLC task. In addition, we want to analyze the relationships between propaganda techniques to understand whether some techniques share common traits, which could be helpful for the classification and tagging tasks.


\bibliography{emnlp-ijcnlp-2019}
\bibliographystyle{acl_natbib}

\appendix


\end{document}